\documentclass[conference]{IEEEtran}
\IEEEoverridecommandlockouts
\usepackage{cite}
\usepackage{amsmath,amssymb,amsfonts}

\usepackage{algorithm}
\usepackage{subfigure}
\usepackage{multirow}
\usepackage{array}
\usepackage{booktabs}
\usepackage{stfloats}
\usepackage{bbding}
\usepackage{hyperref}
\usepackage{algorithmic}
\usepackage{graphicx}
\usepackage{textcomp}
\usepackage{xcolor}
\def\BibTeX{{\rm B\kern-.05em{\sc i\kern-.025em b}\kern-.08em
    T\kern-.1667em\lower.7ex\hbox{E}\kern-.125emX}}
    
\begin{document}

\title{A Reinforcement Learning Benchmark for Autonomous Driving in Intersection Scenarios\\

\thanks{This work was supported in part by the National Natural Science Foundation
of China (NSFC) under Grants No. 61803371, and also was supported by the
Beijing Science and Technology Plan under Grants Z191100007419002 and
the Beijing Municipal Natural Science Foundation under Grants L191002}
}

\author{\IEEEauthorblockN{Yuqi Liu\textsuperscript{1,2}, {\text{Qichao Zhang}\textsuperscript{1,2}}, {\text{Dongbin Zhao}\textsuperscript{1,2}}, \text{IEEE Fellow}}
\IEEEauthorblockA{
\textit{1.The State Key Laboratory of Management and Control for Complex Systems, Institute of Automation,
Chinese Academy of Sciences} \\
\textit{2.School of Artificial Intelligence, University of Chinese Academy of Sciences}\\
Beijing, China \\
\url{liuyuqi2018@ia.ac.cn}, \url{zhangqichao2014@ia.ac.cn}, \url{Dongbin Zhao@ia.ac.cn}}

}

\maketitle

\begin{abstract}

In recent years, control under urban intersection scenarios becomes an emerging research topic. In such scenarios, the autonomous vehicle confronts complicated situations since it must deal with the interaction with social vehicles timely while obeying the traffic rules. Generally, the autonomous vehicle is supposed to avoid collisions while pursuing better efficiency. The existing work fails to provide a framework that emphasizes the integrity of the scenarios while being able to deploy and test reinforcement learning(RL) methods. Specifically, we propose a benchmark for training and testing RL-based autonomous driving agents in complex intersection scenarios, which is called RL-CIS. Then, a set of baselines are deployed consists of various algorithms. The test benchmark and baselines are to provide a fair and comprehensive training and testing platform for the study of RL for autonomous driving in the intersection scenario, advancing the progress of RL-based methods for intersection autonomous driving control. The code of our proposed framework can be found at
\url{https://github.com/liuyuqi123/ComplexUrbanScenarios}.

\end{abstract}


\section{Introduction}

Driving under urban scenarios, especially the intersection scenario, is one of the most challenging problems for an autonomous driving(AD) system. According to\cite{yurtsever2020survey}, a general AD system is composed of several subsystems, including sensing, navigation, decision-making, planning, and control. The key challenge of the intersection scenario is the interaction between the autonomous driving vehicle(ADV) and social vehicles, which mainly possess challenges on the decision-making and control modules\cite{kiran2002deep}. Since the behavioral intention of social vehicles is uncertain, the ADV is forced to negotiate and make decisions quickly under strong interactions, otherwise, traffic accidents are very likely to occur. Generally, the intersection scenario discussed in this paper mainly refers to the ADV passing through a cross-junction while interacting with social vehicles.

Reinforcement learning(RL) methods learn an optimal policy through trial-and-error, a RL agent is able to promote its performance by updating the policy repeatedly. In recent years, RL has become a key technology in the field of AD decision-making and control\cite{wang2019adaptive,zhao2014full,zhao2017model}. In\cite{li2019reinforcement}, deep learning (DL) and RL method is proposed to address the lane-keeping task with the visual input for an ADV on a highway track. And\cite{wang2019lane}, an RL with rule-based constraints method is proposed to train the ADV for the lane-changing task in the highway scenario. 

Besides, some RL-based methods are proposed to deal with the decision-making and control problems in intersection scenarios. In \cite{chen2021interpretable}, an end-to-end framework is proposed for ADV controlling, the RL agent takes raw data as input and directly outputs the control command of the vehicle. The proposed scenario is mixed up with junctions and roundabouts, while the traffic flow in the scenario is not adjustable. In\cite{Bernhard2020}, a multi-agent RL framework is proposed for the behavior model design under intersection scenarios, in which both rule-based and RL-based baselines are provided. The proposed simulator is similar to the one proposed in\cite{highway-env}, both of them fail to provide a delicate dynamic model of vehicles and a high-resolution simulator. In\cite{cai2020summit}, a simulator based on CARLA simulator\cite{Dosovitskiy17} is proposed, which provides a set of real-world road maps as intersection AD benchmark. Though a partially observable Markov decision process(POMDP) agent is proposed, the RL interface is not deployed for further research. In\cite{elsayed20_ultra}, an RL framework named after ULTRA is proposed with a delicate scenario design. In this paper, the behavior of social vehicles are modeled by the SMARTS simulator\cite{zhou2020smarts}. The dynamics of the simulation are idealized as well. We compare the features of some existing frameworks as shown in TABLE \ref{previous work}.

As discussed above, previous work has not been able to integrate a tunable intersection scenario benchmark with the RL-based baseline in a high-resolution simulator. In this paper, we propose a training and testing RL framework addressing the complex intersection scenarios for the AD problem. We call our benchmark Reinforcement Learning Complex Intersection Scenario or RL-CIS for short. As for the original contributions, this paper:
\begin{itemize}
\item proposes benchmark called RL-CIS for RL-based AD agents evaluation, the RL-CIS includes both stochastic and deterministic tests and training environment for RL agent,
\item develops an RL training environment for intersection scenarios and proposes a traffic flow generation method based on stochastic process,
\item provides some basic metrics of performance for the AD agent, also a set of baselines of both RL and rule-based methods. The experimental results show that our RL agent outperforms some rule-based methods in intersection scenarios.
\end{itemize}

\linespread{1.2}
\begin{table*}[hb]
    \caption{Comparison on Framework of Intersection Autonomous Driving}
    \label{previous work}
    \centering
    \begin{tabular}{c|c c c c c}
        \toprule[2pt]
        Features & Vehicle dynamics & RL Interface & Scenario diversity & Scenario randomness & Evaluation benchmark\\
        \hline 
        CARLA scenario runner & \XSolid & \Checkmark & \XSolid & \Checkmark & \Checkmark\\
        Highway env & \XSolid & \Checkmark & \XSolid & \Checkmark & \Checkmark\\
        interp e2e driving & \Checkmark & \Checkmark & \XSolid & \XSolid & \XSolid\\
        Summit & \Checkmark & \XSolid & \Checkmark & \XSolid & \XSolid\\
        BARK & \XSolid & \Checkmark & \XSolid & \Checkmark & \Checkmark\\
        ULTRA & \XSolid & \Checkmark & \Checkmark & \Checkmark & \Checkmark\\
        \textbf{RL-CIS}(ours) & \CheckmarkBold & \CheckmarkBold & \CheckmarkBold & \CheckmarkBold & \CheckmarkBold\\
        \bottomrule[2pt]
    \end{tabular}
\end{table*}

\section{Intersection Scenarios}

\subsection{Design of Test Scenarios}

In order to evaluate the AD agent, we design a set of intersection scenarios based on \cite{groupstandard}, which is proposed for ADV field test evaluation. Inspired by \cite{groupstandard} and\cite{deelman-fgcs-2015}, a scenario is defined through a 3-level structure. The first level is the definition of the functional scenario. In this part, the road network structure, target route, behavior of the traffic flow, and other necessary features are defined. The second level is the logical scenario. During this phase of scenario development, the range of all parameters is given, which constitutes a set of all available scenarios. Then, the final stage is that the concrete scenario, a specific group of parameters is selected to instantiate a single concrete scenario, which will be rendered for the ADV test. 

\subsubsection{Variability}

The variability is the most critical feature for the proposed benchmark. The variability is composed of the diversity of the functional scenarios and the randomness of the logical scenarios, which are the two major rules in scenario design. 

\paragraph{Diversity}

Diversity is emphasized for functional scenarios design. In this phase, the diversity of functional scenarios refers to the variety of the driving tasks and behavior of social vehicles, which leads to a diverse interaction between ego and the social vehicles. In\cite{chen2021interpretable}and\cite{cai2020summit}, though the traffic flow is random, the diversity of interaction between ADV and social vehicles is not guaranteed, which means some type of interaction may not happen among the whole test set.

Specifically, in a cross-intersection, the ADV has 3 potential passing directions, which are turning left, turning right, and going straight. For each driving task, the ego vehicle may encounter interacting traffic flow from different directions. The variation of the interaction is the major diversity of intersection scenarios.

\paragraph{Randomness}

Randomness is the critical rule in logical scenarios design. In intersection scenarios, randomness mainly indicates two aspects. The first is the randomness of the behavior model of social vehicles. It is defined as that how a social vehicle reacts to potential traffic conflicts. The second aspect is the range of critical kinetic parameters, such as the target speed and minimum brake distance for some rule-based methods. In the design of the range of tunable parameters, there is a trade-off. The parameter range should be as broad as possible to generalize the scene, but it should not exceed a reasonable range.

\begin{figure}[hp]
    \centering
    \includegraphics[width=8cm]{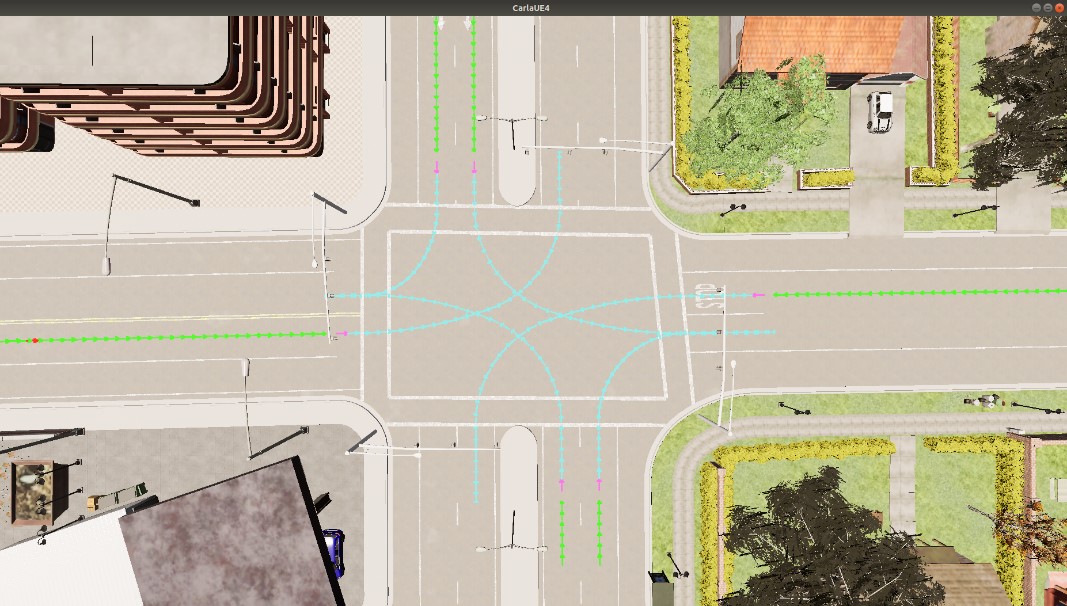}
    \caption{Intersection scenario in CARLA simulator}
    \label{carla intersection}
\end{figure}

\subsubsection{Deterministic Test Scenarios}

The un-signalized intersection scenario is one of the most challenging scenarios for ADV and is mainly discussed in this paper. In the signalized intersection, the interaction between vehicles is taken over by the traffic lights assuming that all traffic participants obey the traffic rules. Though in real-world traffic, both signalized and un-signalized junctions are common. The un-signalized intersection is selected intentionally to stress the interaction problem in the intersection scenario.

In this paper, we select the cross-intersection scene which is the most typical road-network structure among intersection scenarios. We use the Town03 map in the CARLA simulator to configure the scenario, as shown inf \ref{carla intersection}. The blue lines refer to the planned turning routes in the intersection. In a cross-intersection, three typical tasks are mainly considered, which are turning left, turning right, and going straight. For each turning task, there are multiple potential interacting traffic flows which is deterministic by the road network structure. To further decompose scenarios, each functional scenario extracts only a single interacting traffic flow. 

\begin{figure}[htbp]
    \centering
    \subfigure[Turning left facing a going straight traffic flow]{
        \includegraphics[width=3.9cm]{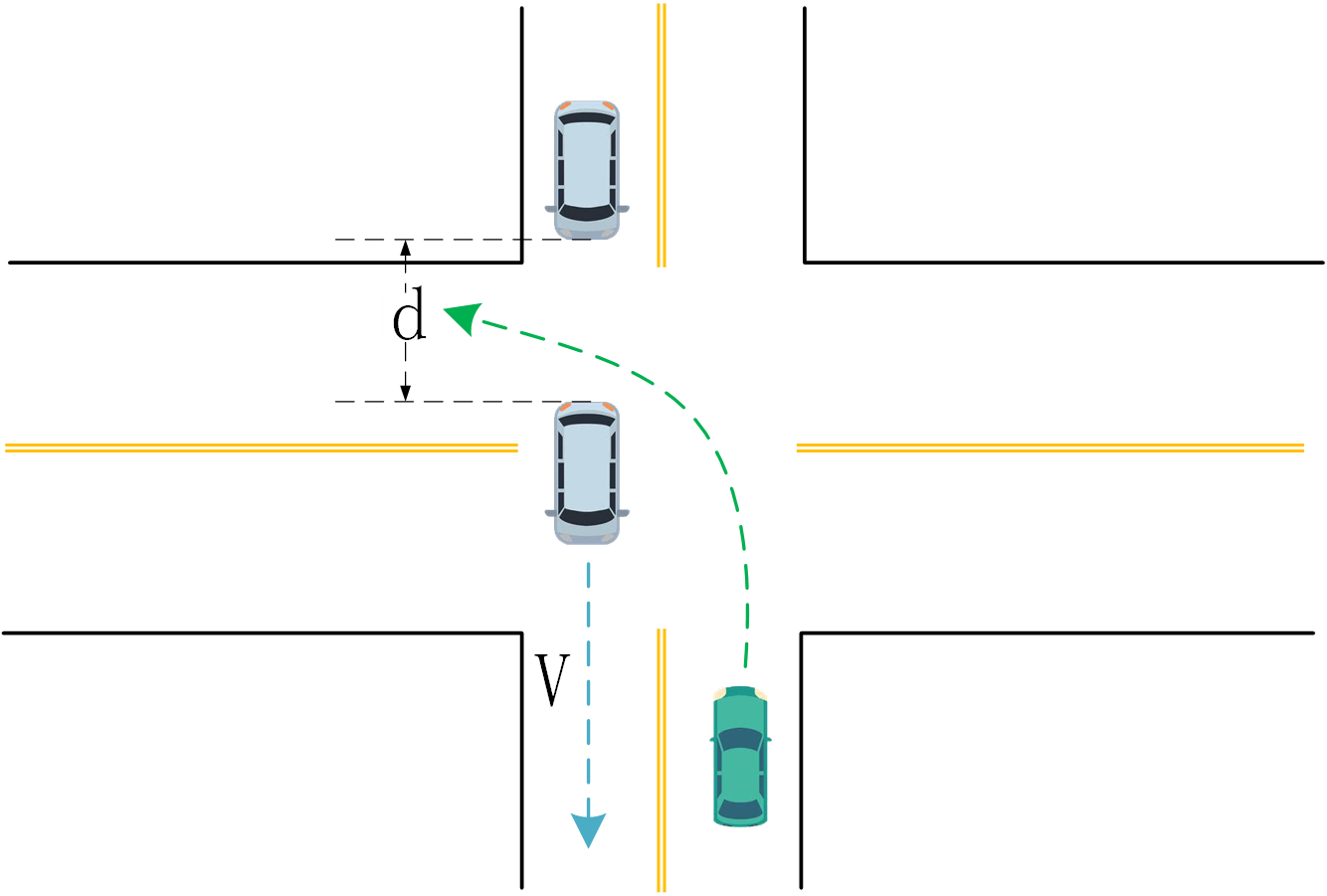}
    }
    \quad
    \subfigure[Turning left facing a right turning traffic flow]{
        \includegraphics[width=3.9cm]{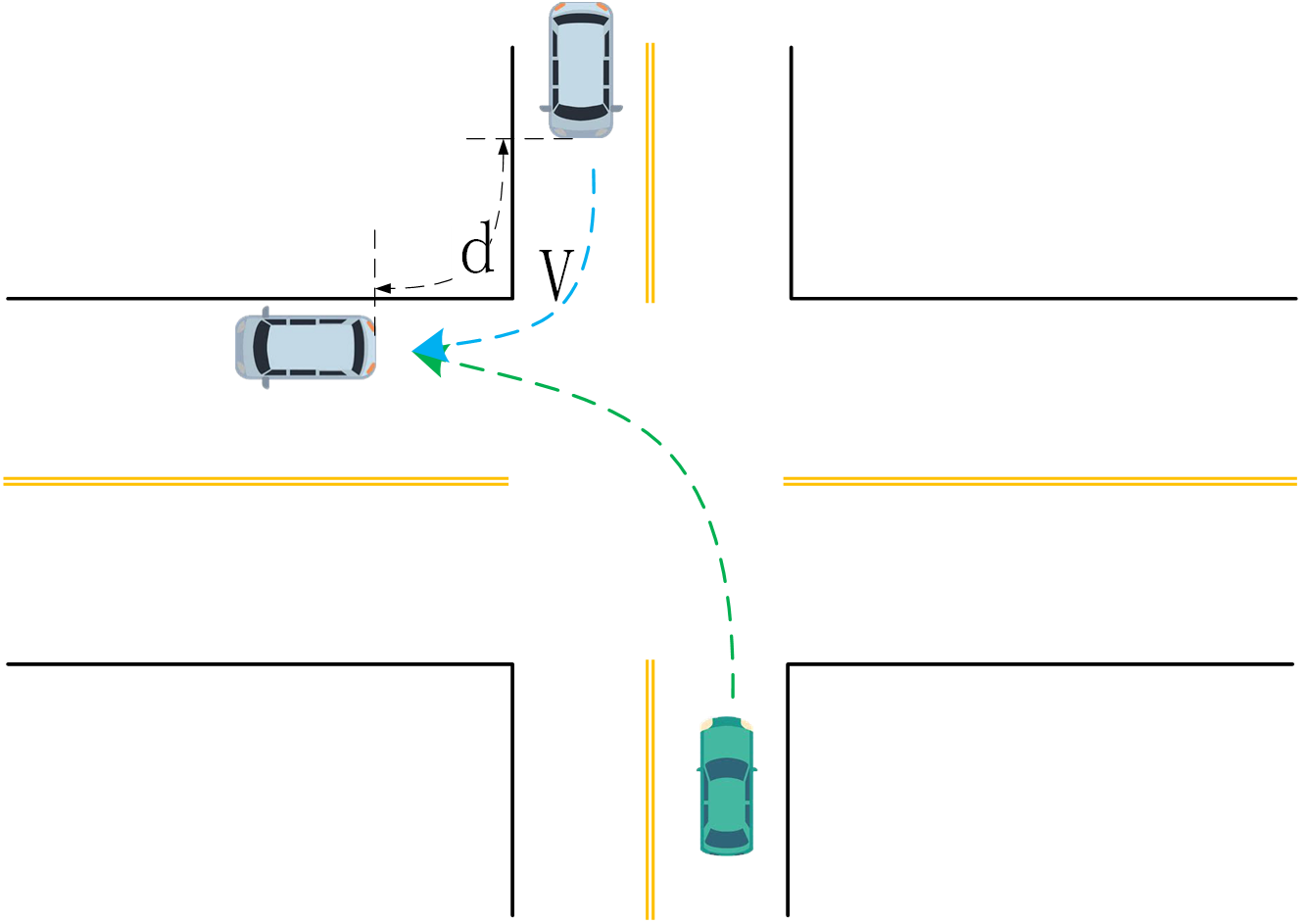}
    }
    
    \subfigure[Turning right facing a going straight traffic flow]{
        \includegraphics[width=3.9cm]{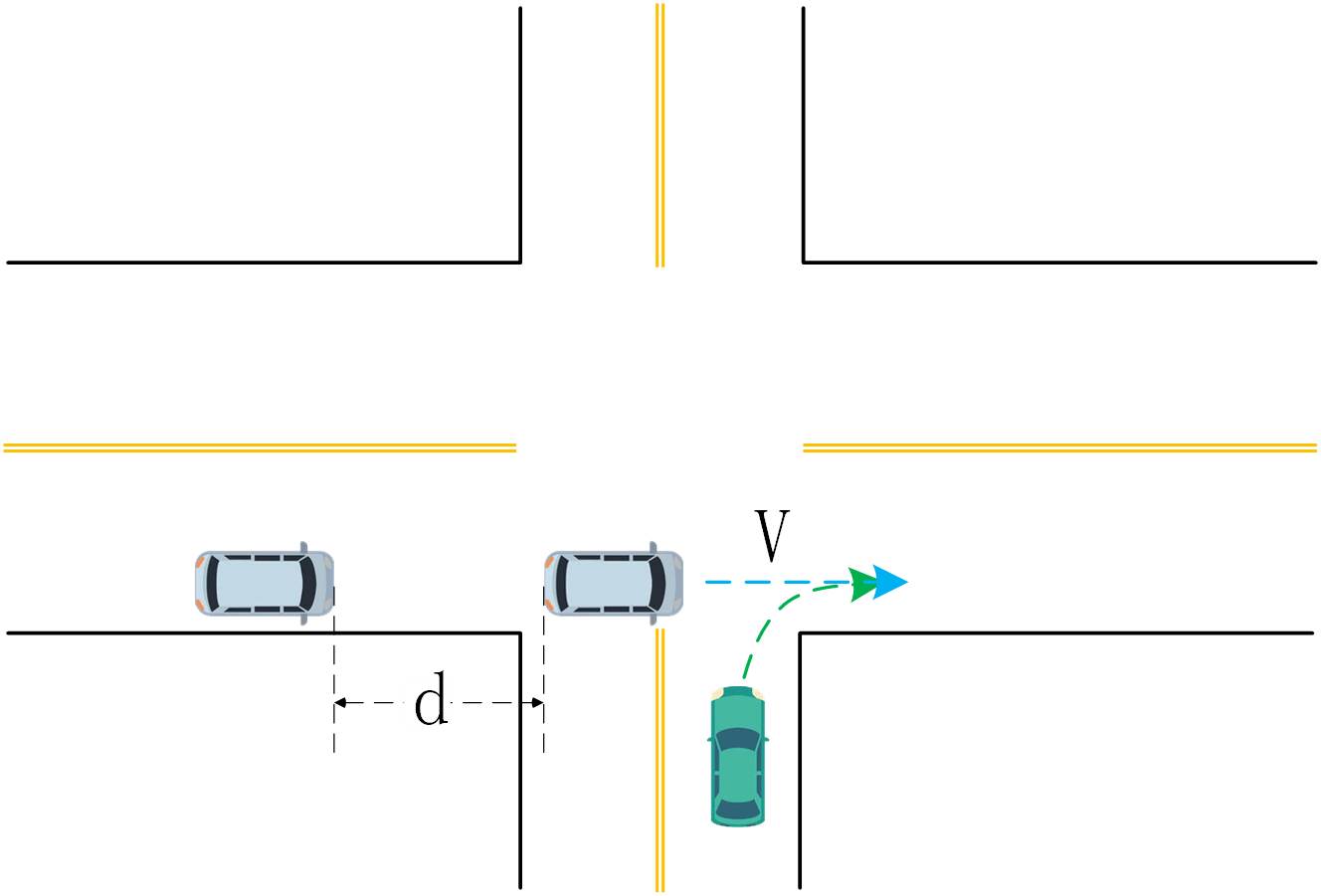}
    }
    \quad
    \subfigure[Going straight facing a going straight traffic flow]{
        \includegraphics[width=3.9cm]{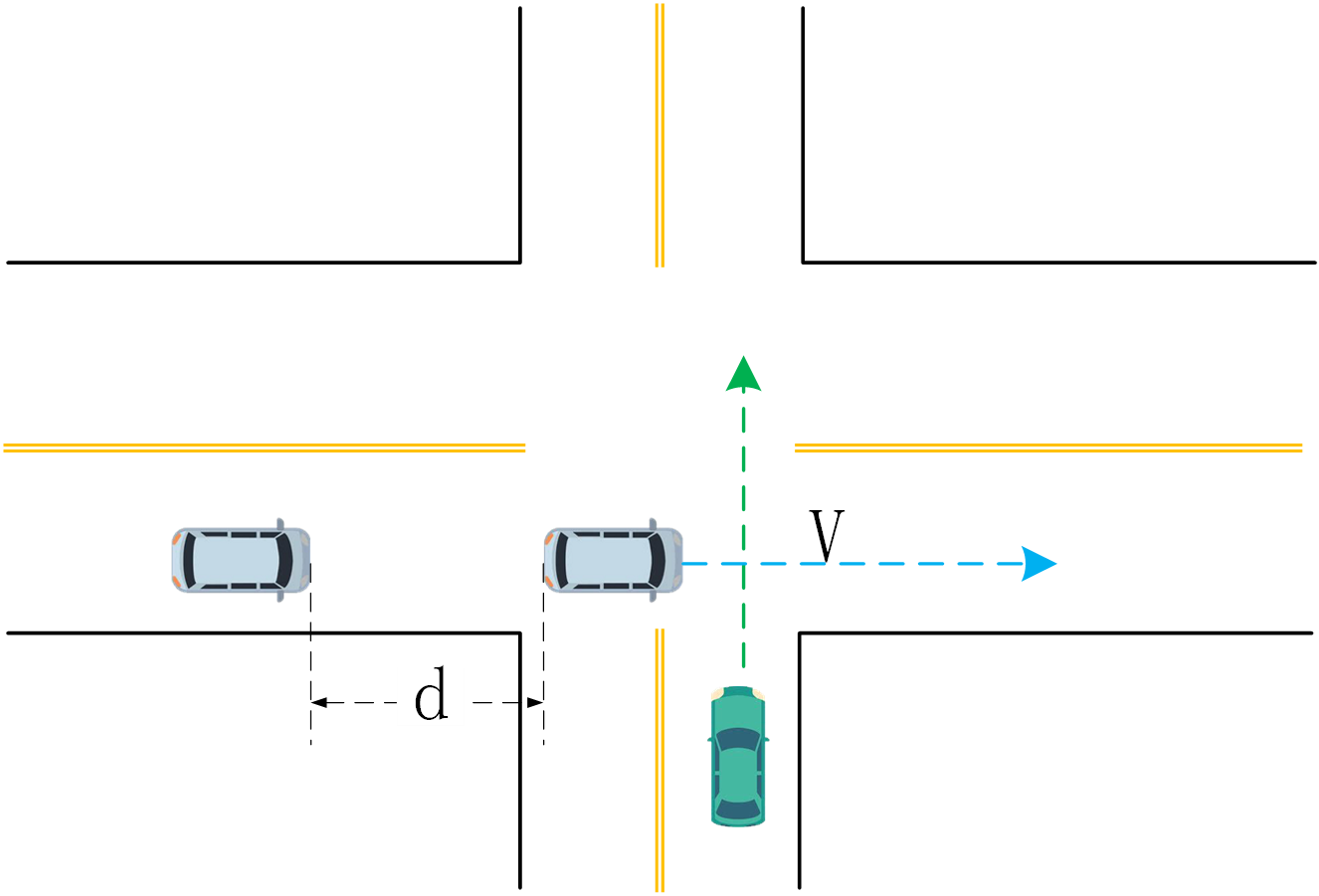}
    }
    
    \subfigure[Going straight facing a left turning traffic flow]{
        \includegraphics[width=3.9cm]{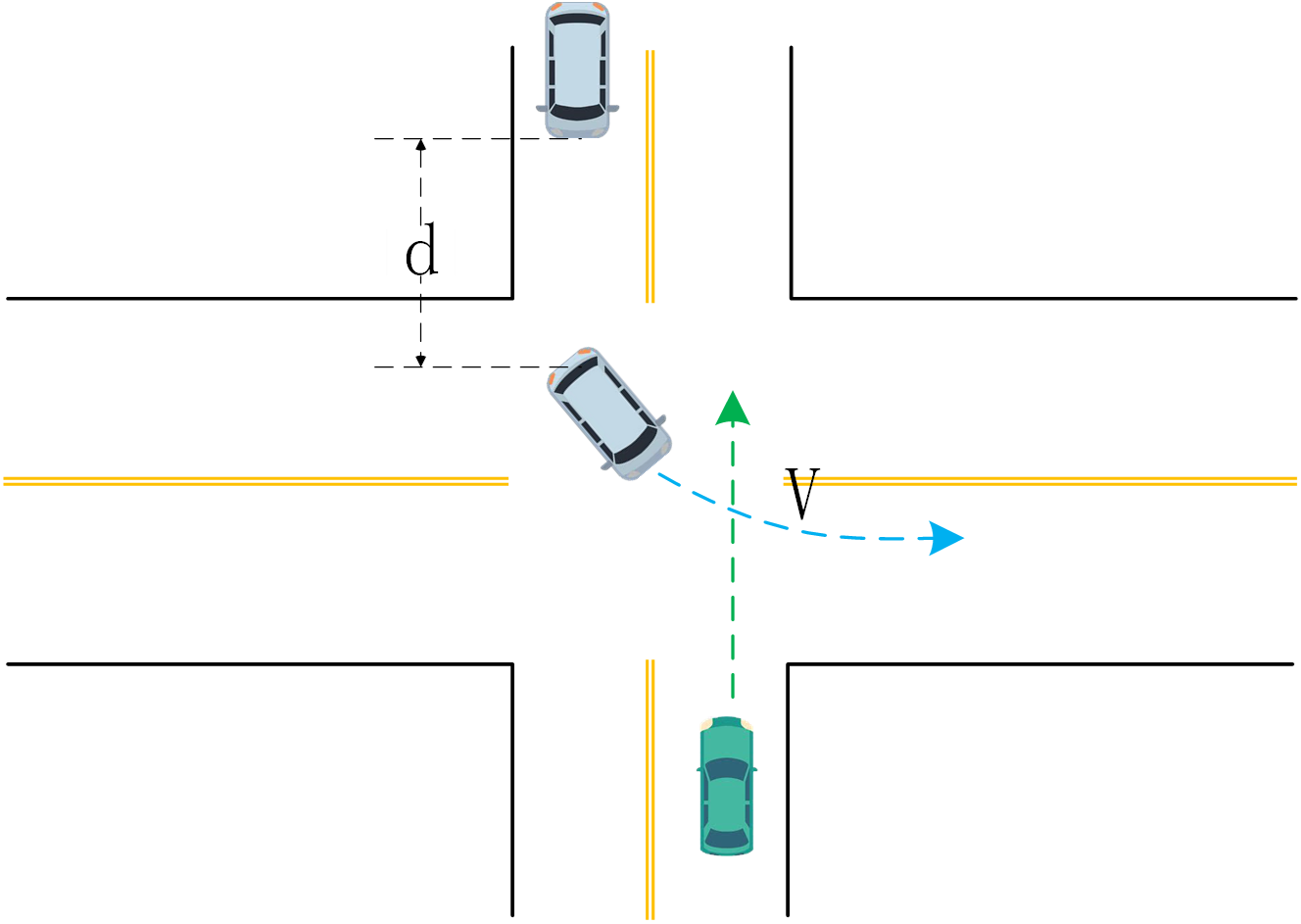}
    }
    
\caption{Functional scenarios of intersection scene}
\label{intersection_scenario}
\end{figure}

Therefore, we propose five functional scenarios as shown in Fig.~\ref{intersection_scenario}. The functional scenario (a) and (b) refer to left-turning task, interacting with a going straight and turning right traffic flow coming from the opposite direction. Secondly, scenarios (c) and (d) refer to ego vehicle going straight task, interacting with a going straight and turning left traffic flow coming from left and opposite direction respectively. Thirdly, scenario (e) refers to turning right task interacting with a going straight traffic flow coming from the left direction. The traffic flow in each scenario is composed of a continuous sequence of social vehicles, each vehicle in the traffic flow has a predefined route as illustrated in Fig.\ref{carla intersection}. 

The behavior model of all social vehicles is determined by two rules. The first one is speed tracking. The social vehicle will accelerate until it reaches the target speed and then maintain it. The second rule specifies how the social vehicle reacts to the potential conflict, which we employ the autonomous emergency braking(AEB) method. Generally, the AEB model detects a certain range in the front direction, if any obstacles are detected, the vehicle will brake until the collision detection is clear, the vehicle will continue pursuing the target speed.

Besides the design of functional scenarios, the logical scenario is instantiated with a set of determined kinetic parameters. In our proposed intersection scenarios, the adjustable kinetic parameters are the target speed of each vehicle in the traffic flow $V$ and the gap distance between adjacent vehicles $d$, as shown in \ref{intersection_scenario}. The gap distance is fixed in each concrete scenario instance to guarantee the stability of the test. The two parameters are determined through discretization of a certain range. The value of target velocity is sampled from $[10, 40]$km/h uniformly with a step length of 2. The value of gap distance is sampled from $[16, 50]$m uniformly with a step length of 2.

\subsubsection{Stochastic Test Scenarios}

Besides the deterministic test, we propose the stochastic test set for the RL-based agent evaluation. In the stochastic test, a comparatively more random traffic flow is provided. Specifically, the behavior model of social vehicles is determined by CARLA's built-in Autopilot function. CARLA Autopilot is a rule-based AD framework which includes navigation, planning and control module. The social vehicle controlled by CARLA Autopilot will randomly plan the route to pass through the junction. For the driving behavior, a CARLA autopilot agent has a Boolean switch for collision avoidance with certain vehicles. In our experiment, we switch off the collision avoidance of all social vehicles against RL-based ADV. Because the RL agent learns the policy through trial-and-error, switching off the collision detection will help the RL agent explore and learn policies more efficiently.

\subsection{Design of Training Scenarios}

Generally, the training scenario set is supposed to cover the test scenario set as much as possible but remains some variability at the same time. The traffic flow setting in training scenarios is similar to the one used in test scenarios. We deploy a rule-based traffic flow with adjustable kinetic parameters, while each traffic flow holds a fixed route. The kinetic parameters are target speed and gap distance, which are the same as in the deterministic test. However, in the training scenario, the two parameters are various for each vehicle of the same traffic flow. The parameters are sampled from the same interval as defined in the deterministic test when a new social vehicle is spawned. The behavior model of all social vehicles in training scenarios uses the same assumption as in the deterministic test, which combines speed tracking and the AEB model.

Therefore, we deploy three agents for turning left, turning right, and going straight tasks respectively. In each training procedure, the functional scenario with the same task route is trained at the same time. For example, in the left turning task, the RL agent will confront traffic flows shown in Fig.\ref{intersection_scenario} (a) and (b) at the same time. The traffic flow on other roads will not be activated. 

\subsubsection{OU-process Parameter Generation}

In the training phase, the RL agent is supposed to fully explore all available traffic situations. As discussed above, the randomness of scenarios is highly affected by the distribution of the kinetic parameters of the social vehicles, which are target speed and gap distance in our proposed test scenarios. Therefore, we deploy a kinetic parameter generation method based on Ornstein–Uhlenbeck(OU) process\cite{finch2004ornstein}. The OU process will generate a sequence of kinetic parameters for the whole traffic flow, and each social vehicle of the traffic flow will be given a set of parameters sequentially.

For the kinetic parameters generation, the OU process has two significant advantages. The first is that the cumulative probability distribution of the OU process is Gaussian. The second is that the OU process is mean reverting, which limits the difference between two contiguous sampling values. The stochastic differential equation(SDE) of the OU process is 

\begin{equation}
d{V_t} = \theta \left( {\mu  - {V_t}} \right)dt + \sigma d{W_t}
\label{ou_sde}
\end{equation}

\noindent in which $\mu,\sigma>0$, $\mu$ refers to the expectation of the variable, $\theta>0$ denotes the damping factor of the OU process, $V_t$ refers to the target speed of the social vehicle, $W_t$ denotes the Wiener process. The ordinary differential equation(ODE) of the OU process is

\begin{equation}
{V_{t + \tau }} = \left( {1 - {e^{ - \theta \tau }}} \right)\mu + {V_t}{e^{ - \theta \tau }} + \sigma \int_t^{t + \tau } {{e^{ - \theta (\tau  - s)}}} d{W_s}
\label{ou_ode}
\end{equation}

Since the kinetic parameters of vehicles is bounded by an interval, inspired by\cite{huang2014limit} we deploy a clipped OU process to avoid over-accumulation on the interval boundary. The clipping process is shown as Algorithm \ref{alg1}. We denote the interval for parameter sampling as $\left[v_l ,v_u \right]$.

\begin{algorithm}
  \caption{Clipped OU process for traffic flow parameter generation}
  \label{alg1}
  \begin{algorithmic}
  \STATE Initialize ${{\hat V}_{t + \tau }}$ with equation \eqref{ou_ode}
  \WHILE{${{\hat V}_{t + \tau }} \notin \left[v_l ,v_u \right]$}
    \STATE Sample ${\hat V}_{t + \tau }$ from OU process using equation \eqref{ou_ode}
  \ENDWHILE
  \STATE $V_{t + \tau }={\hat V}_{t + \tau }$ as the kinetic parameter for next vehicle
  \end{algorithmic}
\end{algorithm}

As for the gap distance between social vehicles, the value is sampled from a truncated gaussian distribution

\begin{equation}
f(d ; \mu, \sigma, d_l, d_u)=\left\{
    \begin{array}{rcl}
    & 0 , & {d<d_l}\\
    & \frac{1}{\sigma} \frac{\phi\left(\frac{d-\hat{V}}{\sigma}\right)}{\phi\left(\frac{d_u-\hat{V}}{\sigma}\right)-\phi\left(\frac{d_l-\hat{V}}{\sigma}\right)}, & {d_u<d<d_l}\\
    & 0, & {d>d_u}\\
    \end{array} \right. 
\end{equation}

\noindent in which $\phi$ refers to the normal distribution, $d_l, d_u$ refers to the value interval of the gap distance, $\sigma$ denotes the variance of the truncated gaussian distribution. In this paper is determined through

\begin{equation}
\sigma = \frac{d_u - d_l}{n}
\end{equation}

\noindent where $n>0$ is a tunable parameter. It is used to adjust the concentration of the parameters relative to the mean value. According to our proposed sampling method, the gap distance is a linear mapping of target speed over the sampling interval. 


\section{Baseline Methods}

\subsection{Baselines for Intersection Scenarios}

\subsubsection{Rule-based Methods}

In this part, we select several classic rule-based methods and an RL algorithm as our proposed baseline methods for the intersection scenarios.

\paragraph{Intelligent Driver Model(IDM)}

The intelligent driver model(IDM)\cite{kesting2010enhanced} is one of the most popular rule-based baselines for the ADV. The IDM model is designed based on the car-following behavior. The IDM model is defined by the following equations

\begin{equation}
\begin{aligned}
&\dot{x}_{\alpha}=\frac{\mathrm{d} x_{\alpha}}{\mathrm{d} t}=v_{\alpha} \\
&\dot{v}_{\alpha}=\frac{\mathrm{d} v_{\alpha}}{\mathrm{d} t}=a\left(1-\left(\frac{v_{\alpha}}{v_{0}}\right)^{\delta}-\left(\frac{s^{*}\left(v_{\alpha}, \Delta v_{\alpha}\right)}{s_{\alpha}}\right)^{2}\right) \\
&\text { with } s^{*}\left(v_{\alpha}, \Delta v_{\alpha}\right)=s_{0}+v_{\alpha} T+\frac{v_{\alpha} \Delta v_{\alpha}}{2 \sqrt{a b}}
\label{IDM_eq_1}
\end{aligned}
\end{equation}

\noindent where $v_{0}$ refers to the desired velocity that the vehicle would drive at in free traffic, $s_{0}$ refers to the minimum desired net distance to the car in the front, $T$ refers to the minimum possible time to the vehicle in front, $a$ refers to the maximum vehicle acceleration, $b>0$ refers to a comfortable braking deceleration, $\delta$ usually takes a value of 4.

Besides, the acceleration of vehicle $\alpha$ can be separated into a free road term and an interaction term

\begin{equation}
    \begin{aligned}
        \dot{v}_{\alpha}^{\mathrm{free}} & =a\left(1-\left(\frac{v_{\alpha}}{v_{0}}\right)^{\delta}\right) \\
        \quad \dot{v}_{\alpha}^{\mathrm{int}} & =-a\left(\frac{s^{*}\left(v_{\alpha}, \Delta v_{\alpha}\right)}{s_{\alpha}}\right)^{2} \\
        & =-a\left(\frac{s_{0}+v_{\alpha} T}{s_{\alpha}}+\frac{v_{\alpha} \Delta v_{\alpha}}{2 \sqrt{a b} s_{\alpha}}\right)^{2}
    \end{aligned}
    \label{IDM_eq_2}
\end{equation}

\paragraph{Autonomous Emergency Braking(AEB) Model}
The AEB method is a widely used technique of level-2 ADV. In real-world deployment, the AEB method processes sensor data with a determined algorithm and performs the brake action if any collisions are detected. In the CARLA simulator, we directly use the ground-truth value as the input for the deployment of the AEB method.

In our work, the AEB method is determined with two rules. The first one is that the longitudinal range of detection $L$ and the second one is the expansion factor of the bounding box of the social vehicle $\eta$. During the driving process, the ego vehicle will detect along its longitudinal direction with a length of $L$. In meantime, the bounding box of each social vehicle is expanded from its original physical model according to the expansion factor. More specifically, the size of the bounding box is calculated through

\begin{equation}
    L_x = \eta \hat{L_x}, L_y = \eta \hat{L_y}, L_z = \eta \hat{L_z}
\end{equation}

\noindent where $L_x, L_y, L_z$ refer to the original size of social vehicle physical model. If the bounding box of any social vehicle penetrating the front area of the ego vehicle, the AEB model will perform a maximum braking action until the detection area is clear.

\subsubsection{Reinforcement Learning}

\paragraph{State Representation}

For the intersection scenario, inspired by \cite{leurent2019social}, we use the ground-truth value of kinetic information of ego vehicle and social vehicles for state representation. Such consideration is common in autonomous driving system design since the major challenge of the intersection scenario comes from the interaction between ego vehicle and social vehicles. For the ego vehicle, the state vector is defined as $s_e = [v_e, g]$, in which $v_e$ denotes the speed of ego vehicle, and $g$ denotes a 3-dimension one-hot vector, which indicates ego vehicle’s current position. For the social vehicle, the state vector of each one is defined as $[v_{i, x}, v_{i, y}, x_i, y_i, cos(\theta_i), sin(\theta_i)]$, in which $v_{i, x}, v_{i, y}$ refer to two-dimensional velocity of social vehicle $i$, $x_i, y_i$ indicate vehicle's Cartesian coordinates and $\theta$ denotes the heading angles under the ego vehicle's coordinate system. The total state representation is combined of the ego vehicle and 5 nearest social vehicles. All six state vectors are concatenated to a 33-dimension vector as the RL input state vector.

\paragraph{Action Space}

The action space is constructed as a 2-dimension continuous variables $a=[a_{0}, a_{1}]$. The vector is transformed for speed tracking by $ \hat a = a_{0} - a_{1}$. Then we scale the action $\hat a$ to $[0,9]$m/s as the target speed of ego vehicle for longitudinal control.

\paragraph{Reward Design}

Inspired by \cite{tram2018learning}, the reward function is defined through events. More specifically, the reward function is combined with two parts, the first is the reward for each timestep, the second is the final reward at the end of an episode. The complete reward function can be written as follows

\begin{equation}
    r=\left\{
    \begin{array}{rcl}
    -0.1 & , & {t \leq 0.5 \cdot T_{max} }\\
    +150 & , & {success}\\
    -350 & , & {collision}\\
    -150 & , & {time \ exceed}\\
    \end{array} \right. 
\end{equation}

\noindent where $T_{max}$ refers to the maximum time limit of one episode, by which common sub-task reward encourages ego vehicle to improve traffic efficiency.

\paragraph{RL Algorithms}

For the intersection scenario, we deploy the TD3 algorithm\cite{fujimoto2018addressing} as our RL baseline. For the neural network design, the state vector is divided by ego and social vehicle, each component is followed by an encoder network, which is formed by $64\times64$ fully-connected(FC) layers. The output of encoders is concatenated and followed by an FC layer. The actor and critic network of the TD3 algorithm share the same network structure in our experiments.

\section{Simulation Experiments}

\subsection{Evaluation Metrics}

\subsubsection{Intersection Scenarios}

Many metrics can be used to measure the behavior of agents \cite{tampuu2020survey}. For an AD system, safety and efficiency are the most concerned performance index. In our framework, success rate and average passing time are general metrics for performance evaluation. Success rate indicates that how the RL agent performs for the specified task directly. In this paper, the success rate is defined as 

\begin{equation}
Success Rate = \frac{Success Counts}{Total Test Number} \times 100 \%
\end{equation}

Secondly, efficiency is measured through the average duration time of a single testing episode. Since the functional scenario which targets on same turning task shares the same route, the average passing time is compared through route classification. It is important to note that we only count the time of the successful test. That is to say, scenarios are divided into three groups for average passing time comparison.

\subsection{Results and Analysis}

\subsubsection{Training Process}

\begin{figure}[hp]
    \centering
    \subfigure[Reward]{
        \includegraphics[width=3.9cm]{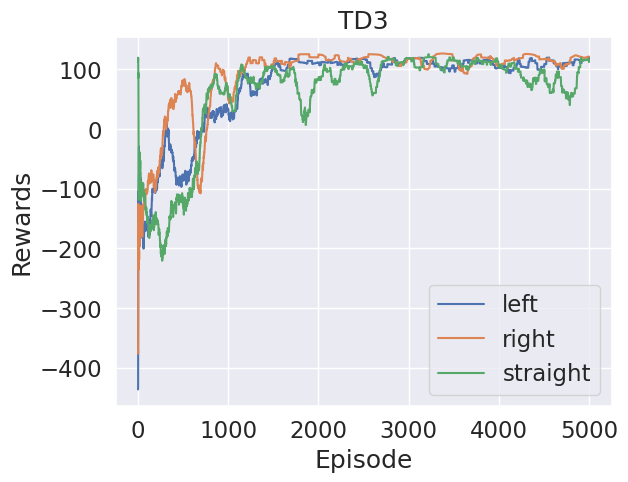}
    }
    \quad
    \subfigure[Success rate]{
        \includegraphics[width=3.9cm]{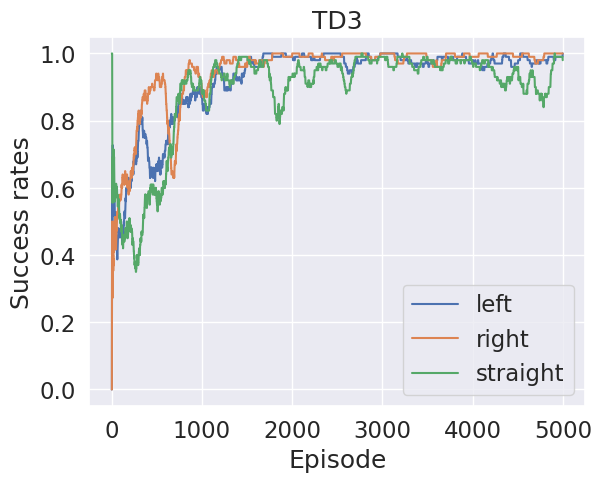}
    }
    \caption{Intersection training curve}
    \label{intersection training curve}
\end{figure}

The learning curves of RL baselines in the task of highway training scenario are depicted in Fig.~\ref{intersection training curve}. From the figure, we can see that the RL agent for each task converges fast within 2000 episodes. Besides, for each route task, the learning curve converges to a maximum level within 5000 episodes. Then the left turning and right turning agents maintain relatively high and stable performance while the going straight agent has slight fluctuation. The main reason for such appearance is mainly because that the agent in going straight task must interact with the left turning traffic flow coming from the opposite direction, which brings greater challenge than other tasks.

\begin{figure}[h]
    \centering
    \subfigure[Going straight traffic flow in left turning task]{\includegraphics[width=8cm]{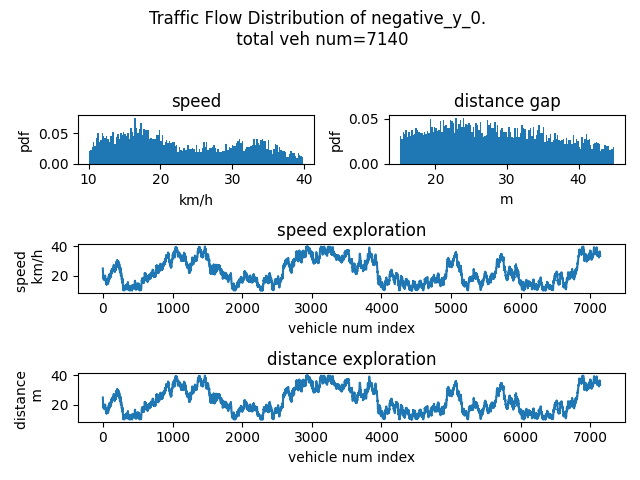}}
    \quad
    \subfigure[Turning right traffic flow in left turning task]{\includegraphics[width=8cm]{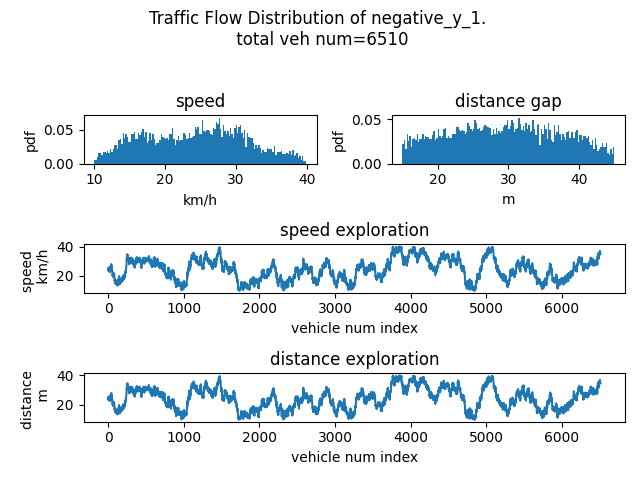}}
    \caption{OU process-based kinetic parameters sampling and distribution}
    \label{ou result}
\end{figure}

In the training process of each task, the interacting traffic flows are activated according to the definition of logical scenarios, as described in experiment settings. The kinetic parameters generation of each traffic flow is counted, the sampling and distribution in the training of left-turning task is shown in Fig.\ref{ou result}. The distribution of parameters sampling is approximate to Gaussian distribution, while the sampling curve of the time domain is rather smooth.

\subsubsection{Deterministic Test}


In the deterministic test, we deploy a TD3 agent as the RL baseline. Besides, two rule-based methods are deployed as a comparison, which is IDM and the AEB model. In our deployment in CARLA, the IDM model will detect a certain distance along the driving route, pursuing either speed tracking or car following behavior as shown in (\ref{IDM_eq_1}) and (\ref{IDM_eq_2}). Compared to the AEB method, the IDM agent adjusts its velocity rather smoothly. The experimental results are shown in TABLE \ref{intersection results}. We evaluate the TD3 agent and rule-based agents in all five functional scenarios. Since the rule-based agents are poorly performed relatively. The statistics are calculated by the task routes for the rules-based agents. In turning left and turning right experiments, the RL agent reaches a success near 90\%, and exceeds the rules-based agent in both success rate and average time. According to the result, going straight is the most challenging task since the left-turning social vehicles are rather fast and hard to negotiate for the ego vehicle.

\linespread{1.2}
\begin{table}[h]
\caption{Intersection Deterministic Test Results. }
\label{intersection results}
\begin{center}
    \begin{tabular}{c c|c|c|c|c|c}
        \toprule[2pt]
        \multicolumn{2}{c|}{Functional scenario} & (a) & (b) & (c) & (d) & (e) \\
        \hline
        \multicolumn{2}{c|}{Ego route} & \multicolumn{2}{c|}{Turning left} & \multicolumn{1}{c|}{Turning right} & \multicolumn{2}{c}{Going straight} \\
        
        \hline
        \multirow{2}{*}{TD3}
         & Success rate(\%) & 94.8 & 93.8 & 89.24 & 99.0 & 80.0 \\
         & Average time(s) & 6.83 & 6.65 & 7.04 & 6.94 & 4.79 \\
        
         
        
        \hline
        \multirow{2}{*}{IDM} & Success rate(\%) & \multicolumn{2}{c|}{67.7} & \multicolumn{1}{c|}{62.8} & \multicolumn{2}{c}{47.4} \\
         & Average time(s) & \multicolumn{2}{c|}{8.40} & \multicolumn{1}{c|}{8.23} & \multicolumn{2}{c}{8.59} \\
        
        \hline
        \multirow{2}{*}{AEB} & success rate(\%) & \multicolumn{2}{c|}{72.74} & \multicolumn{1}{c|}{50.0} & \multicolumn{2}{c}{48.96} \\ 
         & average time(s) & \multicolumn{2}{c|}{8.66} & \multicolumn{1}{c|}{7.21} & \multicolumn{2}{c}{7.18} \\
        
        \bottomrule[2pt]
    \end{tabular}
\end{center}
\end{table}

\subsubsection{Stochastic Test}

The experimental results of the stochastic test are shown in Table~\ref{intersection stochastic test results}. In the stochastic test, the kinetic parameters of traffic flows are uniformly sampled from an interval, which makes traffic flows not extremely dense. Therefore the RL agent reaches a significantly higher success rate for each task compared to the deterministic test.

In this part, the RL method is compared with rule-based methods as well. As the table shows, the RL agent outperforms rule-based methods in both success rate and average time. The overall success rate of the TD3 agent is above 90\%. Though in turning left task and going straight task, rule-base agents have better the average time, they are at a definite disadvantage in terms of safety. We believe such results occur because the IDM and AEB methods have a limited input, which makes them not capable of detecting potential conflict with social vehicles from the cross direction. In conclusion, the model-free RL agent dominates our proposed benchmark of the intersection scene.

\linespread{1.2}
\begin{table}[h]
    \caption{Intersection Stochastic Test Results}
    \label{intersection stochastic test results}
    \centering
    \begin{tabular}{c c|c c }
        \toprule[2pt]
        Methods & Driving task & Success rate(\%)$\uparrow$ & Average time(s)$\downarrow$ \\
        \hline
        \multirow{3}{*}{TD3(ours)}
        & Left & \textbf{95.9} & 9.78 \\
        & Right & \textbf{96.9} & \textbf{7.64} \\
        & Straight & \textbf{91.9} & 9.20 \\
        \hline
        
        \multirow{3}{*}{IDM}
        & Left & 68.3 & 11.21 \\
        & Right & 72.7  & 11.05 \\
        & Straight & 33.7 & 22.6 \\
        \hline
        
        \multirow{3}{*}{AEB}
        & Left & 71.3 & \textbf{9.06} \\
        & Right & 88.7 & 8.86 \\
        & Straight & 58.3  & \textbf{8.83} \\
        
        
        \bottomrule[2pt]
    \end{tabular}
\end{table}

\section{Conclusion}

In this paper, we propose RL-CIS as a framework to train and test the RL-based AD agent in intersection scenarios. Firstly, a group of un-signalized intersection functional scenarios is designed. Then the behavioral model of social vehicles is determined with two critical parameters, composed of the whole logical scenario set. The concrete scenarios are proposed by discretizing logical scenarios and become the deterministic test set in our proposed framework. Besides, we deploy a set of stochastic tests to further evaluate the RL-based AD agent. Meanwhile, the training environment for the RL agent is developed. In this part, a stochastic process-based sampling method is deployed for traffic flow parameters generation. Both the training and test sets are built through the CARLA simulator. In addition to that, we offer a set of baselines for the intersection AD benchmarks, including TD3, IDM, and AEB methods. According to the experimental results, the RL agent shows significant superiority compared with rule-based methods.


\bibliographystyle{IEEEtran}

\bibliography{intersection_ref}

\end{document}